\documentclass{article}

\usepackage{arxiv}
\usepackage[utf8]{inputenc}
\usepackage[T1]{fontenc}
\usepackage{amsmath,amsfonts}
\usepackage{graphicx}
\usepackage{float}
\usepackage{booktabs}
\usepackage{siunitx}
\usepackage{caption}
\usepackage{subcaption}
\usepackage{url}
\usepackage{cite}

\title{Smart Fault Detection in Nanosatellite Electrical Power System}

\author{
  Alireza Rezaee \\
  Department of Mechatronics, School of Intelligent Systems\\
  College of Interdisciplinary Science and Technology\\
  University of Tehran\\
  Tehran, Iran\\
  \texttt{arrezaee@ut.ac.ir} \\
  \And
  Niloofar Nobahari \\
  Department of Mechatronics, School of Intelligent Systems\\
  College of Interdisciplinary Science and Technology\\
  University of Tehran\\
  Tehran, Iran\\
  \texttt{nobahari.niloofar@ut.ac.ir} \\
  \And
  Amin Asgarifar \\
  Faculty of New Sciences and Technologies\\
  University of Tehran, Tehran, Iran\\
  \And
  Farshid Hajati \\
  School of Science and Technology\\
  Faculty of Science, Agriculture, Business and Law\\
  University of New England\\
  Armidale, NSW 2350, Australia\\
  \texttt{fhajati@une.edu.au} \\
}

\begin{document}
\maketitle

\begin{abstract}
This paper presents a new detection method of faults at Nanosatellites’ electrical power without an Attitude Determination Control Subsystem (ADCS) at the LEO orbit. Each part of this system is at risk of fault due to pressure tolerance, launcher pressure, and environmental circumstances. Common faults are line to line fault and open circuit for the photovoltaic subsystem, short circuit and open circuit IGBT at DC to DC converter, and regulator fault of the ground battery. The system is simulated without fault based on a neural network using solar radiation and solar panel’s surface temperature as input data and current and load as outputs. Finally, using the neural network classifier, different faults are diagnosed by pattern and type of fault. For fault classification, other machine learning methods are also used, such as PCA classification, decision tree, and KNN.
\end{abstract}


\section{Introduction}

Nanosatellites, according to their missions, have several subsystems. The power subsystem generates, manages, and distributes power. The power subsystem is essential in all satellites. According to the mission type and requirements, this subsystem includes parts like photovoltaic solar arrays for converting the sun’s energy and generating electrical power, DC to DC Maximum Power Point Tracker (MPPT) converter and regulator for power optimization and regulation. DC charge manager subsystem, loads, and switch mechanism are designed and implemented for a specific purpose \cite{Patel2004}.

The power generated by the photovoltaic solar array and the whole system is affected by the orbital condition and satellite circumstances such as orbital high, inclination, radiant intensity, heat flux, or radiation at orbit temperature that should be analyzed \cite{GonzalezLlorente2013}. When the whole system and components work correctly, the satellite is able to carry on the mission. However, several faults may occur due to the launcher’s shock and pressure. If these faults do not be detected at the right time, they may cause the whole system or mission failure.

\subsection{Related Work}

Beyond satellite power systems, related work on pattern recognition, control, and applied machine learning in power, biomedical, and telecommunication domains provides additional methodological support and motivation for our approach \cite{
AbdoliHajati2014,
Ayatollahi2015,
BarolliAINA2024,
BarolliBWCCA2019,
BarolliWAINA2019,
Barzamini2012,
CremersACCV2014,
Fiorini2019,
Hajati2017Surface,
Hajati2006FaceLocalization,
Hajati2010PoseInvariant,
Hajati2017DynamicTexture,
Mahajan2024_ens,
Pakazad2006FaceDetection,
Sadeghi2024COVID_new,
Shojaiee2014Palmprint,
Sopo2021DeFungi,
Tavakolian2022FastCOVID_new,
Tavakolian2023Readmission_new,
Wang2022SoftwareImpacts_new,
KarimiRezaee2017Helmholtz,
MohamadzadeRezaee2017Antenna,
Ramezani2024Drones,
Rezaee2008GeneticSymbiosis,
Rezaee2010FIR,
Rezaee2017PID,
Rezaee2017Penetrometer,
Rezaee2017MPC,
RezaeeGolpayegani2012,
RezaeePajohesh2016,
Sadeghi2024ECG,
Taghvaee2014Metamaterial,
Tavakolian2022SoftwareImpacts_new,
Gavagsaz2018LoadBalancing,
Rezaee2014FuzzyCloud,
Sarvghad2011ThinkingStyles,
Shahramian2013Leptin,
Shahramian2013Troponin
}. Research has been conducted on detecting, classifying, and diagnosing electrical equipment faults at the satellite power subsystem. Zhao et al.\ \cite{Zhao2015GBSL} proposed a method to detect faults at photovoltaic solar arrays based on Graph-Based Semi-Supervised Leaning (GBSL). By GBSL, they detected faults that an Overcurrent Protection Device (OCPD) could not uncover. Zhao et al.\ \cite{Zhao2012DecisionTree} detected and diagnosed the photovoltaic solar systems’ faults using a decision tree. Mohamed et al.\ \cite{Mohamed2015} also detected photovoltaic solar systems’ faults by neural networks and the genetic algorithm. In a converter proposed in \cite{Freire2014}, there was a method for detecting faults, detecting open circuit power switch faults without using additional hardware, and using the correct control loop in the PWM converter.

In aerospace projects, fault in Li-Ion battery is applied \cite{He2013SOC} because the density of energy is high and does with methods like diagnosis and distinguish fault based on Kalman Sequential Adaptive filter for looking state charge and DE charge of the battery, rate of charge, and de charge, estimate of state charge and DE charge (SOC), and health state (RUH) with use by back up vector machine and collect load information that is connected to it \cite{Sepasi2014,Nuhic2013}.  

The above-mentioned researches do not consider the fault diagnosis of the whole satellite’s power system. Only two studies have discussed the fault diagnosis in the satellite’s power subsystem. In the first research, the fault diagnosis and tolerance have been discussed based on the PCA method for any voltage, current, and temperature faults at sensors. This paper is only concerned about the accumulative fault on the power system’s sensors \cite{Lee2010PCA}. In \cite{Xie2013Bayesian}, each fault probabilities in the satellite’s power system with Bayesian network is defined by knowledge of scientist but for using the Bayesian category without the use of scientist knowledge, fault probilities must be determined or can be estimated by Gaussian equation \cite{Wong2012Hybrid}.

In this paper, the faults of the whole power system, including photovoltaic parts, the DC to DC converter, and the battery are diagnosed and resolved. This paper separate faults using a detective method.

\section{Materials and Methods}

Satellite electrical power system consists of analog and digital circuits that include transistors, trustors, diodes, amplifiers, logic elements, switches, and connections. These parts are manufactured with specific reliabilities. The faulting rate in a component ($\lambda$) is defined as
\[
\lambda = \frac{N_{\text{faulty}}}{N_{\text{total}} \times t}
\]
where $t$ is the outfit time hour unit (h), $N_{\text{faulty}}$ is the number of faulty elements, and $N_{\text{total}}$ is the total number of elements. This rate depends on temperature and the mode of operation. Table~\ref{tab:fault_rate} lists the faulting rate in various components.

\begin{table}[H]
\centering
\caption{Fault rate at temperature 40 $^\circ$C for various components \cite{Isermann2006,Vazquez2008}.}
\label{tab:fault_rate}
\begin{tabular}{lcl}
\toprule
Equipment & $\lambda$ ($10^{-9}$/h) & \\
\midrule
Transistor                & 10--9   & \\
Trustor (power switch family) & 10--9   & \\
Digital integrated circuit & 30      & \\
Logical elements          & 30      & \\
Analog switches           & 2000    & \\
Amplifier                 & 10--9   & \\
Diode                     & 10--9   & \\
Battery Li-Ion            & 10--9   & \\
Solar array               & 10--9   & \\
\bottomrule
\end{tabular}
\end{table}

Using the faulting rate, we can diagnose faults at electrical power system components such as IGBTs, MPPT converters and regulators, solar array, and battery. Open circuit and line-to-line faults in solar array, short circuit and open circuit faults in IGBTs, MPPT converters and regulators, and ground fault in battery are feasible in nanosatellite power systems.

\begin{figure}[H]
\centering
\includegraphics[width=0.7\linewidth]{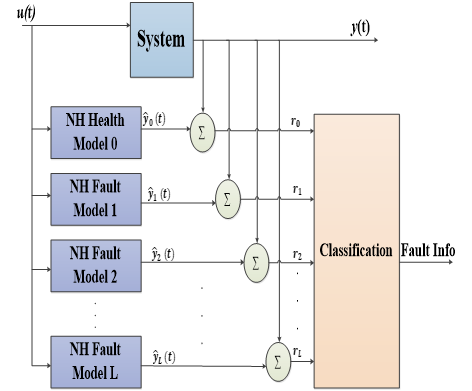}
\caption{Proposed fault diagnosis model.}
\label{fig:block}
\end{figure}

Figure~\ref{fig:block} shows the overall block diagram of the proposed fault diagnosis model. First of all, the system must be recognized by virtual neural network for detecting and classify faults at different systems. At the next step, possible faults intentionally and virtually are applied to the system, then like a normal state, the faulty system is diagnosised. Afterward, for each fault, one model like figure 1 is created that can be compared by system output with output model.

Create remains based on equation (1) then calcify faults:
\begin{equation}
r_i = y(t) - \hat{y}_i(t) \,,
\label{eq:residual}
\end{equation}
where, according to figure 1, system output $y(t)$ is contrasted to output model system without fault $\hat{y}_0(t)$ and output faulty model like $\hat{y}_1(t)$ to $\hat{y}_n(t)$, remained amounts ($r_0$ to $r_n$) according to relation (\ref{eq:residual}) are created and for pattern recognition is categorized with neural network. 

In sequence, at first, the electrical power system is simulated with neural networks without fault which is a powerful tool in system recognition \cite{Patton1991}, and then possible faults are injected into the system and then they are simulated and graph is given.

\subsection{Electrical power system modeling}

The electrical power system of the satellite is provided by a solar array with convert sun radiation power. Sun radiation power of solar array is influenced by environmental circumstances. Because of the array function and power system, both are influenced by the rate of solar radiation and solar cell surface temperature \cite{Vazquez2008}. Solar array temperature is dependent on the environmental circumstance like the rate of sun radiation, thermal flux includes sun thermal flux, albedo thermal flux, and earth sub radiant flux. Solar array productive current is dependent on to rate of solar radiation directly. Therefore, two parameters, sun radiation power and solar array temperature that are directly affected to produce power and satellite power system are system inputs after simulating the power system in MATLAB software.

Nanosatellite has insufficient space, volume, and mass for installing sensors also for decreasing cost of project and satellite launch, for modeling power system is consumed that output load current as output and system according to nonlinear properties as static nonlinear are based on equation (3).
\begin{equation}
I_{\text{load}} = f(P_{\text{sun}}, T) \,,
\label{eq:nonlinear}
\end{equation}
where $I_{\text{load}}$ is the load output current (A), $P_{\text{sun}}$ is sun radiation power (e.g.\ W/m$^2$), and $T$ is solar array temperature (Celsius degree). The electrical power system is consisting of elements and components. Model of each subsystem and create a link between these models is complicated and token a lot of time. Because of this reason for modeling the whole system, one can assume these parts as one part, their inputs and output are considered and the black box is identified by virtual neural networks.

For identifying system from tree layer virtual neural network, with the Marquardt algorithm \cite{Hagan1994}, random data categorizing, elementary activated function, and \texttt{tansig} from mean squire variance of fault are used based on equation (4).
\begin{equation}
E = \frac{1}{n} \sum_{k=1}^{n} \bigl( y_k - \hat{y}_k \bigr)^2 \,,
\label{eq:mse}
\end{equation}
At equation (4), $y_k$ is system output, $\hat{y}_k$ is model output and $n$ is data number. In addition to these, for analyzing neural network accurately is used network output coherent scale, system output and also studies its statistical behavior (average amount and fault variance) from fault graph histogram.

\begin{figure}[H]
\centering
\includegraphics[width=0.9\linewidth]{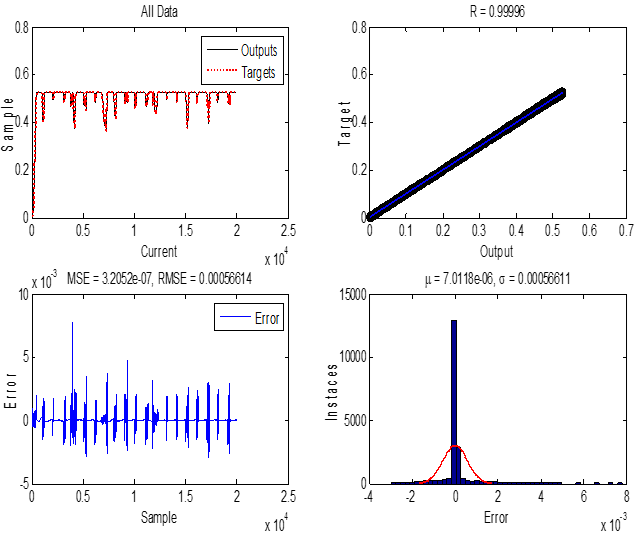}
\caption{System output coherent and its estimated amount, the output electrical power system without fault and its estimated amount diagram, estimated fault histogram and amount of fault square average, and the average of fault square based on samples.}
\label{fig:sys_model}
\end{figure}

Modeling result is shown in Figure~\ref{fig:sys_model}.

\begin{figure}[H]
\centering
\includegraphics[width=0.9\linewidth]{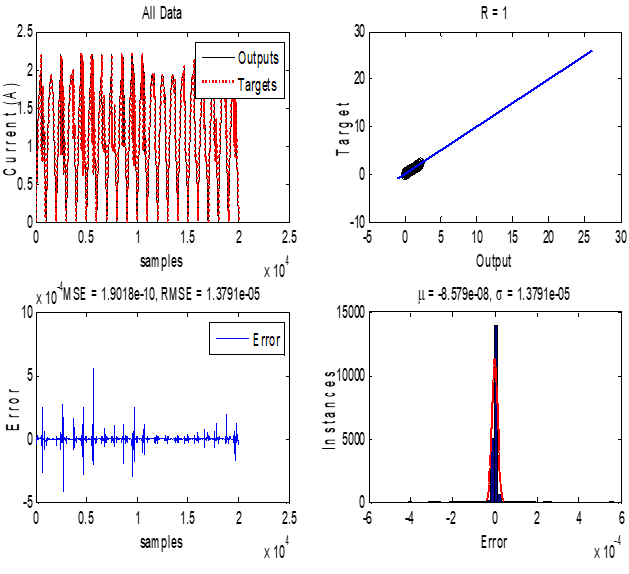}
\caption{Current of photovoltaic system output coherent and its estimated amount, output diagram and its estimated amount, estimated fault histogram, and amount of fault square average and average of fault square based on samples.}
\label{fig:pv_current}
\end{figure}

\begin{figure}[H]
\centering
\includegraphics[width=0.9\linewidth]{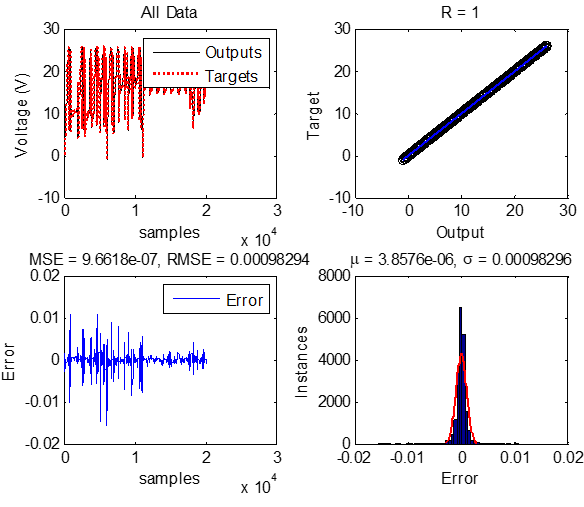}
\caption{Voltage of photovoltaic system output coherent and its estimated amount, output diagram and its estimated amount, estimated fault histogram, and amount of fault square average and an average of fault square based on samples.}
\label{fig:pv_voltage}
\end{figure}

At MPPT converter for calculating power optimization photovoltaic subsystem use current and voltage sensors until to sampling this parameter instantly, that amount will compare with optimization amounts \cite{Subudhi2013}. Because of this for diagnosing and calcification possible faults at photovoltaic subsystem can use these sensors and from moving fault of a solar array at power system prevent and diagnose more quickly. For this reason, for modeling the photovoltaic subsystem as before inputs are sun radiation power and solar array surface temperature and outputs are current and voltage of arrays and with using of neural networks, nonlinear assumption and static, the system is identifying. The results are shown in Figures~\ref{fig:pv_current} and \ref{fig:pv_voltage}.

With a study on Figures~\ref{fig:sys_model} to \ref{fig:pv_voltage} is observed that neural network have high accuracy for output model based on system output is estimated that is oscillating, real and estimate output similarity rate on the system modeling with current load is 96\% and at photovoltaic system model is 100\%. Fault on the modeling is considered parameters as the average of fault square, average and fault variance. At previous simulates, the amount of average square fault is less than $10^{-6}$ at system modeling with load current, at photovoltaic system outputs is less than $10^{-6}$ at the voltage and $10^{-9}$ at current. With a study on faults’ histogram graph at Figures~\ref{fig:sys_model} to \ref{fig:pv_voltage} is observed that faults’ average at system modeling is less than $10^{-5}$, photovoltaic system outputs are less than $10^{-6}$ at the voltage and $10^{-9}$ at current. The fault of variance at previous modeling as the sequence is 0.00056, 0.00013, and 0.00098. At faults’ histogram is seen that the Gaussian estimate of modeling faults at Figures~\ref{fig:sys_model} to \ref{fig:pv_voltage} are ideal at the peak that it is a symptom of faults that aren’t separated from the average amount. As a result, above the neural network has suitable accuracy for the electrical power system model without fault.

\subsection{Faulty electrical power system modeling}

Faults happen as random at electrical and electronic systems \cite{Isermann2006}. With attention to this assumption, each possible fault must inject into the system and identify a system model with a neural network. As a sequence, open circuit and line-line fault at the solar array, MPPT converter IGBT open circuit fault, open circuit fault, short circuit IGBT regulator converter, and battery ground fault are identified based on section 2 and is summarized at Table~\ref{tab:fault_models}.

\begin{table}[H]
\centering
\caption{Summary of faulty electrical power system modeling results.}
\label{tab:fault_models}
\begin{tabular}{llcccc}
\toprule
Subsystem & Fault model & System output coherent & RMSE & Fault average $\mu$ & Fault variance $\delta$ \\
\midrule
Photovoltaic & Open circuit current  & 1          & $10^{-11}$ & $3.6\times 10^{-7}$ & $6.7\times 10^{-6}$ \\
             & Open circuit voltage  & 1          & $10^{-5}$  & 0.00225             & 0.00921 \\
             & Line-line current     & 0.99838    & 0.00056    & 0.00032             & 0.02375 \\
             & Line-line voltage     & 1          & $10^{-8}$  & $10^{-6}$           & 0.00103 \\
IGBT MPPT    & Open circuit load current & 0.99999 & $10^{-8}$  & $5.3\times 10^{-7}$ & 0.00023 \\
IGBT regulator converter & Open circuit load current & 0.99963 & $10^{-10}$ & $10^{-8}$ & $9.5\times 10^{-5}$ \\
                      & Short circuit load current & 0.99986 & $10^{-6}$  & $10^{-6}$           & 0.0122 \\
Battery       & Ground load current  & 1          & $10^{-8}$  & $10^{-8}$           & 0.00012 \\
\bottomrule
\end{tabular}
\end{table}

\subsection{Photovoltaic subsystem faults calcification with neural network MLP}

\begin{figure}[H]
\centering
\includegraphics[width=0.6\linewidth]{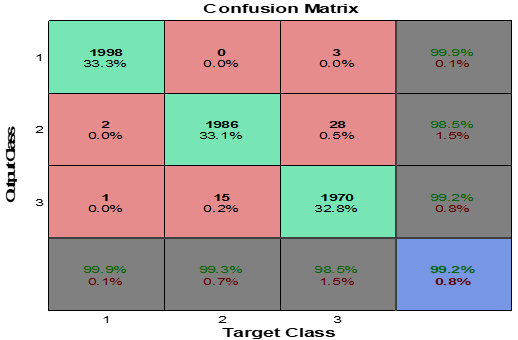}
\caption{Neural network classifier confusion matrix MLP for identifying fault of a photovoltaic system.}
\label{fig:mlp_pv}
\end{figure}

At the photovoltaic subsystem, possible faults are line-line and open circuit. For fault determining and calcification can also use neural network MLP. Healthy photovoltaic system model, line-line, and open circuit fault were calculated in the previous section. Now, the model output vector compare with the system output vector, amount remains are built based on equation (5) and as input enter to classifier.

\begin{equation}
r = \bigl[ V_{\text{pv}} - \hat{V}_{\text{pv}},\ I_{\text{pv}} - \hat{I}_{\text{pv}} \bigr] \,,
\label{eq:pv_residual}
\end{equation}
At equation (5), $r$ is the remaining amounts, $V_{\text{pv}}$ and $I_{\text{pv}}$ are the amounts of voltage and current are for the photovoltaic system, $\hat{V}_{\text{pv}}$ and $\hat{I}_{\text{pv}}$ are the amounts of output voltage and current of a photovoltaic system. The number of simulation data for the classifier is 2001 that which is provided for each class. The output is 3 bit that identifies faults classes. The result of this calcification is shown at the confusion matrix Figure~\ref{fig:mlp_pv}.

A study on Figure~\ref{fig:mlp_pv} is defined that neural network accuracy for classification of possible fault at photovoltaic system is more than 99.9\% that explains its accuracy but calcification accuracy at model classes without fault as the sequence is line-line and open circuit. Also, it is clear that from 2001 model class data without fault 1998 data is classified correctly and 3 data have a fault, it means that the accuracy detects of this class is approximately 99.9\%. In photovoltaic system class accuracy of fault is 99.3\% from 2001 data, 1968 data are correct and 15 data are not correct that are calcified. Also in open circuit fault class accuracy is 98.5\% from 2001 data, 1970 data is correct and 31 data is not correct that is detected. A study on Figure~\ref{fig:mlp_pv} also can see that from 3 data which are not correct and are diagnosed in system class without fault, 2 data are in line–line fault class and 1 data are in open circuit fault class that are classified incorrectly. Classification fault in line-line fault class from each 15 data is in open circuit fault class and from 31 data must be in open circuit fault class. 28 numeric data are in line-line fault class and 3 numeric data are in model class without fault.

\subsection{Electrical power system fault classification with neural network MLP}

\begin{figure}[H]
\centering
\includegraphics[width=0.6\linewidth]{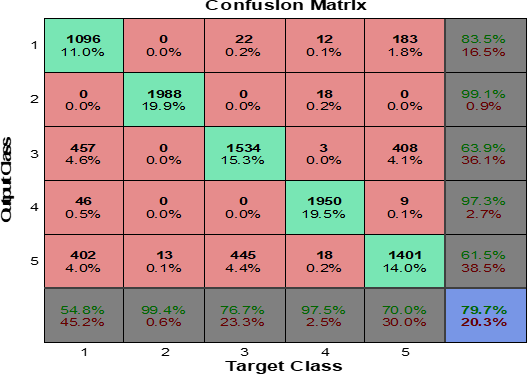}
\caption{Confusion matrix neural network classifier MLP at the first state.}
\label{fig:mlp_all}
\end{figure}

For classification, another possible fault at the electrical power system from the neural network MLP tree layer with allocating 5 bit to 5 output class, consist of a system without fault, the battery is a ground fault, open circuit IGBT converter MPPT fault, open circuit fault, and short circuit IGBT regulate convertor. Inputs of neural clarification are based on figure 1 that are consists of neural models output and output amount of the system based on equation (6).
\begin{equation}
x = [r_0, r_1, \dots, r_4]
\label{eq:mlp_input}
\end{equation}
In equation (6), $r_i$ are remain amounts and $x$ is the input vector to neural classifier figure 1. With enter input of equation (6) to confusion matrix classifier, Figure~\ref{fig:mlp_all} is gained.

Figure~\ref{fig:mlp_all} shows that neural network MLP can diagnose 1096 data from 2001 data at first-class and has 54.8\% accuracy and 45\% fault. As sequence 457, 46, and 402 are in open circuit fault IGBT converter MPPT, open circuit fault and short circuit IGBT regulate convertor. Classifier accuracy at ground battery fault class is more than other classes and is more than 99\%, 13 data in IGBT short circuit fault class is incorrect. Rate of accuracy in the designed neural network at the separation between classes, MPPT converter open circuit fault, and open circuit fault and regulate convertor IGBT short circuit as sequent are 76.7\%, 97.5\%, and 70\% it means that accuracy in separate IGBT open circuit fault regulator converter is good and was diagnosed incorrectly 12 data class without fault, 18 data at ground battery fault, 3 data at IGBT open circuit MPPT converter fault and 18 data at regulating convertor IGBT short circuit fault class. Finally, the confusion matric is shown in Figure~\ref{fig:mlp_all} which this neural network MLP at several states of Nanosatellite electrical power system has 79.7\% accuracy and has 20.3\% fault. Therefore, with these results and accuracies cannot use the network and its outputs for system faults classification.

\begin{figure}[H]
\centering
\includegraphics[width=0.6\linewidth]{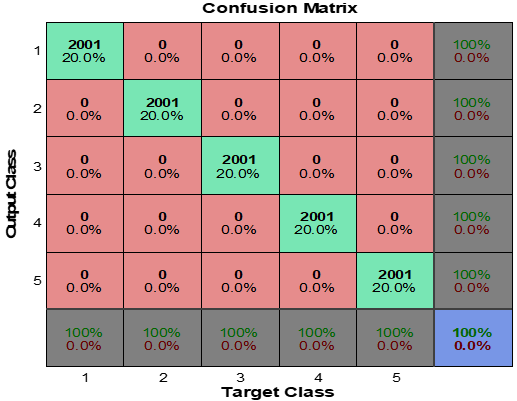}
\caption{Confusion matrix neural network MLP classification with 5 bits’ output.}
\label{fig:mlp_moment}
\end{figure}

For overcoming to insufficient accuracy of the neural network and increase accuracy can use from another feature of the output signal (load current). Adding the first Moment vector (current amount average for every moment) output current as one of the generated properties, based on equation (7) and (8) is identified.
\begin{equation}
m_1 = \frac{1}{N} \sum_{k=1}^{N} I_{\text{load}}(k)
\end{equation}
In equations (7) and (8), $m_1$ is the first output current moment and $r_i$ are remain amounts of neural models. With entering previous outputs to tree layer neural network MLP and study 2001 data at each class of confusion matrix, this classification was calculated according to Figure~\ref{fig:mlp_moment}. Confusion matrix in Figure~\ref{fig:mlp_moment} is diagonal and shows that 100\% of data are correctly categorized in fifth classes and designed neural network MLP could be detected all state of a faulty system with high accuracy because 20\% of data, it means 2001 data is each class and classification fault in each class based on Figure~\ref{fig:mlp_moment} is zero.

\subsection{Fault classification with another intelligence method}

Calculate the average amount of output load current at each moment may impose a heavy calculating load to the system processor, for this reason, can use another intelligence method for separating system faults.

In the satellite electronics power system for controlling and managing battery charge and de charge use SOC parameter with battery charge state. A precision estimate of the battery charge state protects from battery health, unnecessary charge, and de charge of the battery increases the life of the battery, decreases the probability of happening fault, and increases system reliability \cite{Nuhic2013}. SOC parameter is calculated by Kalman. With the use method \cite{He2014SOC} SOC was calculated. Now with use intelligence classification methods like PCA, KNN, and decision tree can proceed to possible fault classification of power system with current load inputs and battery SOC.

\subsubsection{KNN method}

For fault classification also can use the KNN method. KNN is a lazy classifier and data is classifying without build model and learning samples. With the use KNN classifier method and with use output current inputs, battery charge state and build Ordered pair $(I_{\text{load}}, \text{SOC})$ are defined possible fault in satellite electrical power, $K$ is geometric distance Euclidean based on equation (9), around each Ordered pair $(I_{\text{load}}, \text{SOC})$ is used for data classification \cite{Gaddam2007}.
\begin{equation}
d = \sqrt{(I_{\text{load}} - I_k)^2 + (\text{SOC} - \text{SOC}_k)^2}
\end{equation}
In the above equation $d$ is the geometric distance, $I_{\text{load}}$ is loaded current and $(I_k,\text{SOC}_k)$ is point $k$ charge state in the plane of figure (7). For evaluating the function at this algorithm can proceed to study on cross-validation and Resubstituting Loss parameters. In cross-evaluation, data is separated into $K$ group, and then learning operation and cross-test on each group, K-fold Loss parameter is defined.  

\begin{figure}[H]
\centering
\includegraphics[width=0.6\linewidth]{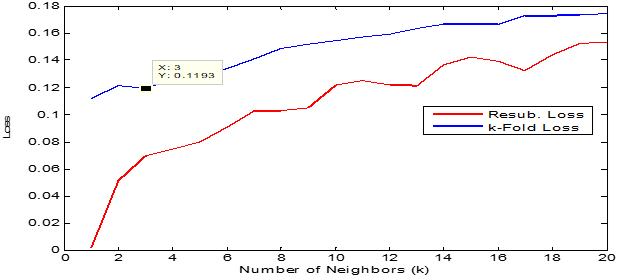}
\caption{K-Fold faults and replacement at KNN method.}
\end{figure}

With a study on figure 8 is determined that number of neighbors in the KNN method must choose equivalent 3 with Resubstituting Loss and K-Fold Loss because Resubstituting Loss fault one or two near neighbor is down but choose $K$ parameter equivalent 1 and 2 cause classifier over coherency to data learning. To determine the number of neighbors, fault classification is done and is presented at Table 3. 
\[
\text{Resubstitution Loss} \approx 0.0698 \sim 7\%, \quad \text{K\_Fold Loss} \approx 0.1193 \sim 12\%
\]

\begin{table}[H]
\centering
\caption{Result of KNN classification method.}
\begin{tabular}{lcc}
\toprule
Class & Accuracy (\%) & Notes \\
\midrule
System without fault & 86 & \\
Battery ground fault & 100 & \\
Open circuit IGBT MPPT & 86 & \\
Open circuit regulator & 99.99 & \\
Short circuit IGBT regulator & 92 & \\
\bottomrule
\end{tabular}
\end{table}

At the table (3) is observing that the KNN algorithm cannot classify fault completely.

\subsubsection{Decision tree}

For separate and apart existence classes can use a learning tree or decision tree. Decision tree is the method for estimating objective with discrete amounts. This method is one of the most famous of a posteriori learning algorithms that use in the different application successfully and this method has resistance to input data noise. Decision tree learning is using the ID3 algorithm. Based on decision tree calcification is choosing a strategy that has the most likelihood \cite{Gaddam2007}.

For separating system with or without fault from the decision tree method is using output load current and battery charge state such as PCA and KNN methods, table (4) is presenting fault calcification result, replacement fault rate and k replicate.
\[
\text{Resubstituting Loss} \approx 0.0137 \sim 1.3\%, \quad \text{K\_Fold Loss} \approx 0.0435 \sim 4.3\%
\]

\begin{table}[H]
\centering
\caption{Decision tree classification result.}
\begin{tabular}{lcc}
\toprule
Class & Accuracy (\%) & Notes \\
\midrule
System without fault & 97 & \\
Battery ground fault & 99.99 & \\
Open circuit IGBT MPPT & 98 & \\
Open circuit regulator & 99.6 & \\
Short circuit IGBT regulator & 98.35 & \\
\bottomrule
\end{tabular}
\end{table}

\subsubsection{PCA method}

PCA method is a linear mapping between inputs and outputs that the discrete system is based on their Eigenvalue and Eigenvector. Figure 9 is presenting inputs the plane of output load current and battery charge state at system fault class and possible species fault sin system at two dimensions, before act PCA method.

\begin{figure}[H]
\centering
\includegraphics[width=0.6\linewidth]{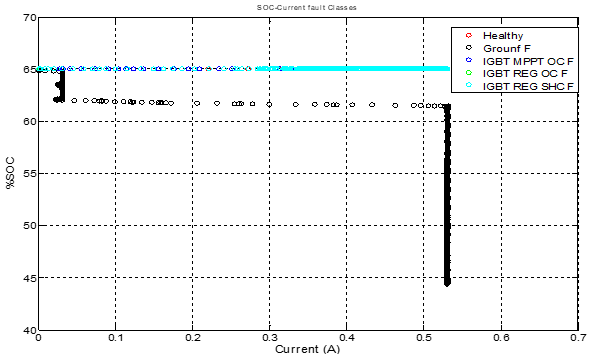}
\caption{Fault species on the plane of battery charge state and load current.}
\end{figure}

Figure 8 is presenting tat states of without faults and possible fault classes on the electrical power system at two dimensions’ plane of battery charge state and output load current cannot discrete.

For using the PCA method, first of all, inputs classifiers are Ordered pair $(I_{\text{load}},\text{SOC})$ and matrix $X$ at equation (9) is considered.
\[
X = \begin{bmatrix}
I_{\text{load},1} & \cdots & I_{\text{load},N} \\
\text{SOC}_1 & \cdots & \text{SOC}_N
\end{bmatrix}
\]
Also in equation (9), $y$ is mapping output vector and $Q$ is the Eigenvector matrix for covariance matrix $C$ in equation (10). At this equation average data (bias) is equal zero, defined as below:
\[
C = \frac{1}{N} X X^\top
\]
For calculating the equation of the PCA method, matrix $Q$, Eigenvector for the biggest Eigenvalue Matrix $C$ is determined ascending in the column of matrix $Q$ and then choose Eigenvectors that have bigger Eigenvalue as a result produce more variance dispersion in $y$ output. Separation and calcification result fault with the PCA method is without fault state and fault class species are presented in Figure 10. 

\begin{figure}[H]
\centering
\includegraphics[width=0.6\linewidth]{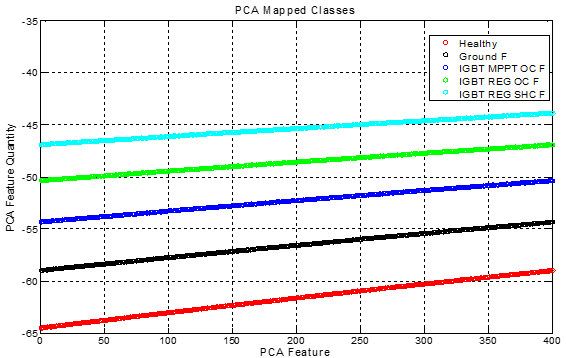}
\caption{Separating possible fault in the electrical power system with the PCA method.}
\end{figure}

Figure 10 is proposing that without fault and possible fault species are reduced to one dimension and with suitable space and are separated by the PCA method.

The result for the conclusion presented methods of separating faults at the electrical power system except for photovoltaic subsystems is brought at the Table~\ref{tab:compare_methods}.

\begin{table}[H]
\centering
\caption{Compare classification methods for electrical power system faults.}
\label{tab:compare_methods}
\begin{tabular}{lcc}
\toprule
Method             & System        & Accuracy (\%) \\
\midrule
Neural network (MLP) & Electrical power & 100 \\
                    & Photovoltaic     & 99.2 \\
PCA                 & Electrical power & 100 \\
KNN                 & Electrical power & 93 \\
DT                  & Electrical power & 98.6 \\
\bottomrule
\end{tabular}
\end{table}

Table (5) depicts that the accuracy of neural network MLP and PCA method which are two algorithms for separate and classification possible fault in the electrical power system of Nanosatellite is more than others. DT method also had high accuracy but the KNN method based on its accuracy cannot use for this purpose. For analysis quality of classification possible fault in the electrical power system in each class, can refer to Table 7 (omitted here for brevity).

\section{Conclusion}

This paper is a proposed simulation of the satellite electrical power system in Nanosatellite. Effective parameters on the function of a power system are inputs and the electrical load current is output because of mass and volume limitation and with consideration to these parameters and use reliability analysis of neural network on system equipment, electrical system model without fault state and with possible fault was determined with suitable accuracy. Then were used classification neural network MLP, PCA, KNN, and DT methods for separating classes and presented that neural network MLP and PCA method can separate and classify without fault system and possible fault classes and were more effective than KNN and DT methods.

\bibliographystyle{unsrt}

\end{document}